\documentclass{article}
\usepackage{spconf,amsmath,graphicx,verbatim}
\usepackage[colorlinks,allcolors=blue]{hyperref}
\usepackage[font=small,skip=0pt]{caption}
\usepackage{enumitem}
\setlist{nosep, leftmargin=14pt}

\usepackage[belowskip=-5pt,aboveskip=0pt]{caption}
\usepackage{mwe} 
\usepackage{multirow}

\usepackage{booktabs}
\usepackage{algorithm,algpseudocode}
\usepackage{blindtext}
\usepackage{hyperref}
\usepackage[table]{xcolor}
\usepackage{graphicx}
\usepackage{amsmath}
\usepackage{booktabs}
\usepackage{overpic}

\usepackage[T1]{fontenc}

\definecolor{mygray}{gray}{.92}

\usepackage{tikz}
\def\checkmark{\tikz\fill[scale=0.4](0,.35) -- (.25,0) -- (1,.7) -- (.25,.15) -- cycle;}

\newcommand{\figref}[1]{Figure \ref{#1}}
\newcommand{\tabref}[1]{Table \ref{#1}}

\title{Spatiotemporal learning with context-aware video tubelets
\newline for ultrasound video analysis}
\name{
\begin{tabular}{c} 
Gary Y. Li$^{*1}$, Li Chen$^{1}$, Bryson Hicks$^{2}$, Nikolai Schnittke$^{2}$, David O. Kessler$^{3}$, Jeffrey Shupp$^{4}$, \\ \it Maria Parker$^{2}$, \it Cristiana Baloescu$^{5}$,  \it Christopher Moore$^{5}$,   \it Cynthia Gregory$^{2}$,  \it Kenton Gregory$^{2}$,  \\ \it Balasundar Raju$^{1}$,  \it Jochen Kruecker$^{1}$, \it Alvin Chen$^{1}$
\end{tabular}}

\address{$^{1}$Philips $^{2}$Oregon Health \& Science University \\ $^{3}$Columbia University Vagelos College of Physicians and Surgeons \\ $^{4}$MedStar Washington Hospital Center   $^{5}$Yale University School of Medicine}

\begin{document}
\maketitle
\begin{abstract}

Computer-aided pathology detection algorithms for video-based imaging modalities must accurately interpret complex spatiotemporal information by integrating findings across multiple frames. Current state-of-the-art methods operate by classifying on video sub-volumes (tubelets), but they often lose global spatial context by focusing only on local regions within detection ROIs. Here we propose a lightweight framework for tubelet-based object detection and video classification that preserves both global spatial context and fine spatiotemporal features. To address the loss of global context, we embed tubelet location, size, and confidence as inputs to the classifier. Additionally, we use ROI-aligned feature maps from a pre-trained detection model, leveraging learned feature representations to increase the receptive field and reduce computational complexity. Our method is efficient, with the spatiotemporal tubelet classifier comprising only 0.4M parameters. We apply our approach to detect and classify lung consolidation and pleural effusion in ultrasound videos. Five-fold cross-validation on 14,804 videos from 828 patients shows our method outperforms previous tubelet-based approaches and is suited for real-time workflows.

\end{abstract}


%
\begin{keywords}
video object recognition, spatiotemporal learning, computer-aided diagnosis, medical ultrasound \footnote{*Corresponding author: \href{mailto:ye.li@philips.com}{ye.li@philips.com}} \footnote{
This study was funded in part by the U.S. Department of Health and Human Services (HHS); Administration for Strategic Preparedness and Response (ASPR); Biomedical Advanced Research and Development Authority (BARDA), under contract number 75A50120C00097. The contract and federal funding are not an endorsement of the study results, product or company.}
\end{keywords}

\begin{figure}[!t]
  \centering
  \includegraphics[width=\linewidth]{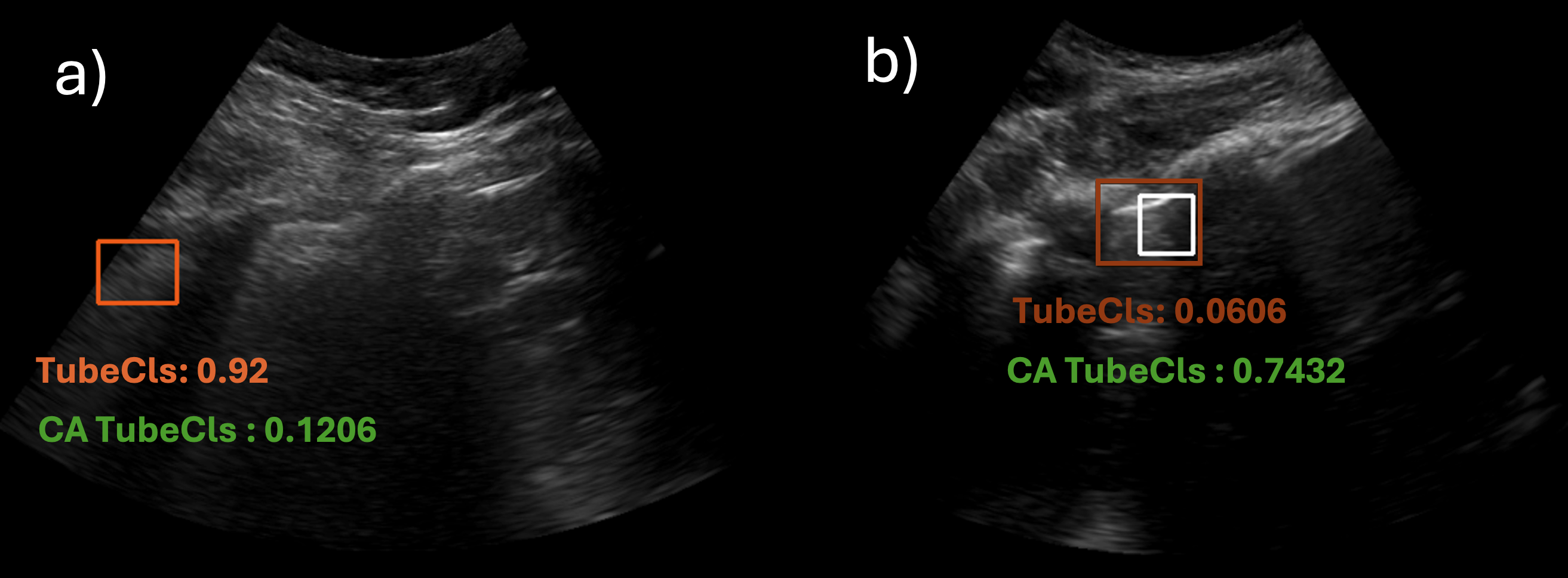}
  \caption{Example ultrasound frames containing pathologies misclassified by tubelet classifier \cite{li2023weakly} but correctly classified by our context-aware tubelet classifier. Orange=tubelets classifier. Green=Context-aware tubelets classifier. White=ground-truth.
  }\label{fig:tubeletClsExample}
\end{figure}

\begin{figure*}[t]
\begin{center}
   \includegraphics[width=\linewidth]{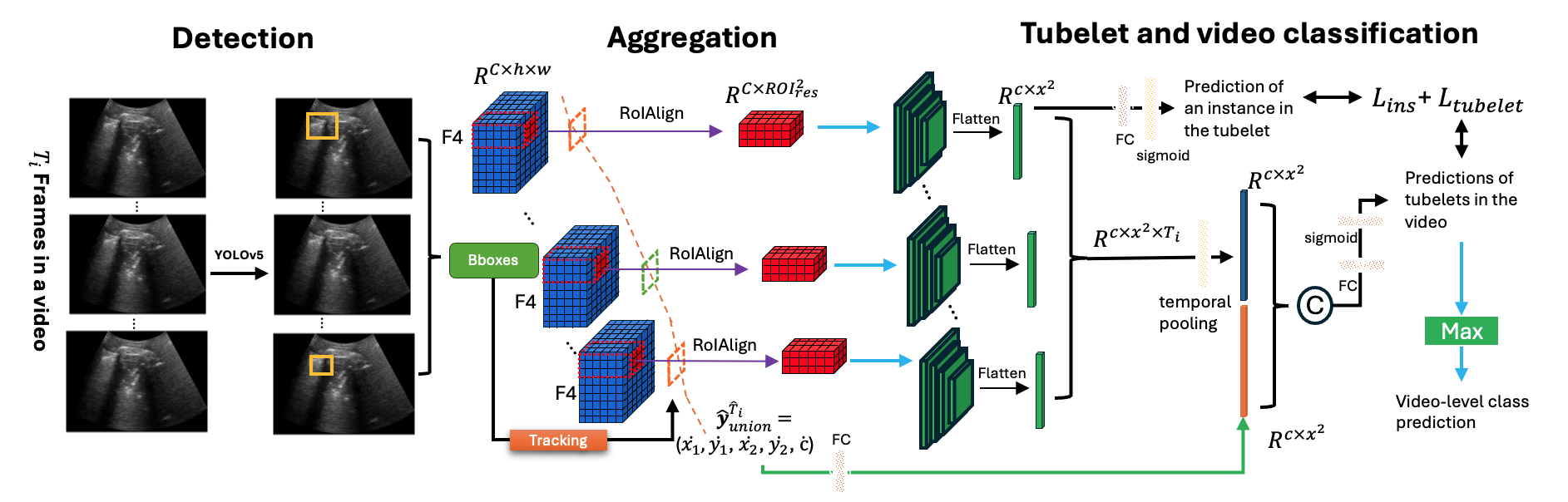}
\end{center}
\caption{Spatiotemporal framework for ultrasound video analysis based on context-aware video tubelets. F4 represents one frame's feature maps from the final convolutional layer of the 4-th stage of the YOLOv5 backbone}
\label{fig:main}
\end{figure*}

\section{Introduction}
\label{sec:intro}


Video-based medical imaging presents unique challenges for developing computer-aided detection due to the need to aggregate complex image data both spatially and temporally for accurate diagnosis. While deep learning excels in video object detection and classification \cite{shea2023deep,dastider2021integrated, sharma2019spatio,hwang2022cannot,li2023weakly,kang2016object,kang2017object}, most models operate as ``black boxes'', obscuring the learned relationship between an object's fine-grained spatiotemporal features and its global context within the scene.

Understanding connections between broader context and finer details is crucial in medical video analysis, where pathology detection depends on the appearance, location, size, and proximity of pathological areas to other key features. For example, in lung ultrasound (LUS), pathologies are confirmed based on relative positions in the image and changes across frames. Features appearing laterally with lower resolution (Fig. \ref{fig:tubeletClsExample}a) or briefly may be dismissed, while central features near anatomical landmarks, like the pleural line (Fig. \ref{fig:tubeletClsExample}b), warrant closer examination. To mimic expert analysis, video classification algorithms must capture both spatial and fine-grained temporal patterns across frames.

To capture finer spatiotemporal features within an ROI, current state-of-the-art solutions \cite{hwang2022cannot, li2023weakly} use a tubelet-based approach, employing a dedicated network to classify a video sub-volume (tubelet). This allows focused learning of detailed spatiotemporal features without background distractions, showing promising performance over end-to-end methods \cite{shea2023deep, dastider2021integrated, sharma2019spatio}. However, these methods lose global spatial context during tubelet classification, focusing only on local regions within detection ROIs. To overcome this, we propose a lightweight framework for tubelet-based object detection and classification that preserves global spatial context alongside fine spatiotemporal features throughout processing. To our knowledge, this is the first approach to introduce spatial context embeddings to retain global information in tubelet-based video classification. Key innovations include:

\begin{enumerate}[label=(\roman*)]
\item \textbf{Embedding global context:}
Tubelet location, size, and detection confidence are embedded as additional inputs in the tubelet classification step. This is the key mechanism to preserve global spatial context after frame aggregation.
\item \textbf{Video tubelets from feature maps:} 
We extract feature maps from intermediate layers of the detector and aggregate them into “video tubelets” based on temporal consistency. The use of feature maps offers two advantages: (a) we capture information outside the detection ROIs, thus increasing spatial context, and (b) we reduce computation by reusing learned features for tubelet classification.
\item \textbf{Lightweight spatiotemporal model:}
We propose a simple and efficient spatiotemporal model for tubelet classification with only 0.4M parameters. The model consists of only 2D convolutions and feature-wise temporal pooling.
\end{enumerate}

\section{Related works}
Recent methods for video classification can be broadly categorized into two approaches: \textbf{1) Direct video classification methods:}
Direct video classifiers based on CNN-LSTMs have been proposed to capture spatiotemporal dependencies in ultrasound \cite{shea2023deep, dastider2021integrated,sharma2019spatio}. 
A challenge with using direct video classifiers, however, is their inability to localize pathology, reducing interpretability. 
\noindent \textbf{2) Frame aggregation methods:}
Alternatively, individual frame predictions can be combined to produce a video-level classification. Frame aggregation can rely on human-designed rules \cite{shea2023deep}, or be learned using temporal models \cite{yue2015beyond,hwang2022cannot}. In \cite{lin2022new}, lesion detection combined clip-level temporal outputs with video-level classification features, while \cite{roy2020deep} used “uninorm” aggregation units to merge frame classifications into video-level COVID-19 severity scores. However, neither method provides locations of identified pathologies. To address the localization task, two recent approaches have been proposed: aggregating detection outputs and aggregating ROI features. (a) \textit{Aggregating detection outputs}: This method forms “tracklets” by linking bounding boxes across frames, enhancing temporal consistency through techniques like high-confidence propagation to neighboring frames \cite{han2016seq}, tracklet re-scoring \cite{kang2016object}, and contextual constraints to reduce false positives \cite{kang2016object}. (b) \textit{Aggregating ROI features}: Feature-based approaches integrate visual and temporal information into “video tubelets.” ROI pooling with an LSTM was used for tubelet classification in \cite{kang2017object}, while \cite{hwang2022cannot} and \cite{li2023weakly} further developed tubelet aggregation using region proposal networks and weakly semi-supervised learning, respectively.

\section{Methods}
\label{sec:typestyle}
\subsection{Spatiotemporal framework for ultrasound video analysis based on context-aware video tubelets}

The proposed framework (\figref{fig:main}) comprises four steps:

\begin{enumerate}[label=(\roman*)]
\item \textbf{Detection:} A pre-trained detector generates bounding box candidates around pathological regions in each frame.
\item \textbf{Aggregation:} Frame detections are aggregated into tracklets based on temporal consistency. Feature maps within the tracklet regions are extracted to form “video tubelets”.
\item \textbf{Tubelet classification with global context:}
Tubelets are processed through a lightweight spatiotemporal model, with tubelet location, size, and detection confidence embedded as additional inputs to provide global context.
\item \textbf{Video classification:}  The final video-level confidence is the maximum of all tubelet predictions in the video.
\end{enumerate}

\subsection{Generating video tubelets from detector features}
\label{gen_video_tubelets}

Instead of constructing video tubelets from original frames as in \cite{li2023weakly}, we extract feature maps from an intermediate detector layer using ROI align \cite{he2017mask}. For an input video $I^{in} \in R^{H \times W \times T} $, we identify high-confidence pathology regions per frame using a pre-trained 2D object detector (YOLO-v5). Detection boxes are then tracked \cite{bewley2016simple} to create tracklets, each represented as $\{\hat{\textbf{y}}^{1},...,\hat{\textbf{y}}^{t}\}$ where $t \in (1,\hat{T}_i)$. Each box vector $\hat{\textbf{y}}$ contains coordinates ($\hat{x_1}, \hat{y_1}$,$\hat{x_2}, \hat{y_2}$) and class confidence $\hat{c}$. To fully preserve the native temporal changes in the ROI, for each tracklet we calculated a set of unified coordinates $\hat{\textbf{y}}_{union}^{\hat{T}_i} = (\dot{x}_1,\dot{y}_1,\dot{x}_2,\dot{y}_2,\dot{c})$, covering the maximum area of all detection boxes, enhancing the temporal consistency and field of view for each instance.

For each tracklet, we used the union coordinates $\hat{\textbf{y}}_{union}^{\hat{T}_i}$ as input to the RoIAlign layer to extract the video tubelet $f_{i}$: 
\begin{equation}
 F_k =concat(F^1_k,...,F^T_k, axis = T) \in R^{C \times h \times w \times T} 
\end{equation}
\begin{equation}
f_{i} = \text{ROIAlign}(F_k,  \hat{\textbf{y}}_{union}^{\hat{T}_i} , \text{ROI}_{\text{res}}) \in R^{C \times \text{ROI}_{\text{res}}^2 \times T_i} 
\end{equation}
\noindent
where $F^i_k \in R^{C \times h \times w}$ is the $k^{th}$ conv. layer’s feature maps across all video frames. $C,h,w$ denote feature map channels, height, and width per frame. 
The output dimension after ROIAlign is denoted by $\text{ROI}_\text{res}$. 
The instance class label for each ROI frame in a tubelet, $L_i^t$ (where $t \in (1,\hat{T}_i)$), is determined by its relation with the frame-level ground truth boxes. 
Given a frame-level bounding box label $\mathbf{y}^{t}=({x}_1^{t}, {y}_1^{t}, {x}_2^{t}, {y}_2^{t})$ and the unified predicted box $\hat{\textbf{y}}_{union}^{\hat{T}_i}=(\dot{x}_1,\dot{y}_1,\dot{x}_2,\dot{y}_2,\dot{c})$ for the same frame, a positive class label $L_i^t=1$ is assigned if the center of the unified box and any ground truth box in that frame fall within each other's extents.

\begin{equation}
\begin{split}
\begin{cases}
    {x}_{1}^{t} \leq \frac{\dot{x}_{1}+\dot{x}_{2}}{2} \leq {x}_{2}^{t}, 
    {y}_{1}^{t} \leq \frac{\dot{y}_{1}+\dot{y}_{2}}{2} \leq {y}_{2}^{t}, \\ 
    \dot{x}_{1}^{t} \leq \frac{{x}_{1}+{x}_{2}}{2} \leq \dot{x}_{2}^{t}, 
    \dot{y}_{1}^{t} \leq \frac{{y}_{1}+{y}_{2}}{2} \leq \dot{y}_{2}^{t}, \\ 
  \end{cases}
\end{split}
\vspace{-5pt}
\end{equation}
\noindent
We assign a label to each video tubelet by aggregating all the instance labels, that is,

\begin{equation}
y_{i} = \begin{cases} 
1, \text{if}~ \sum_{t=1}^{\hat{T}_i}L_i^{t} &> 0, \\
0, \text{otherwise}
\end{cases}
\end{equation}

\subsection{Embedding tubelet location, size, and confidence}
\label{tubelet_embedding}
To address the loss of global context during tubelet classification, we embed the unified tubelet's location, size, and detection confidence from the detector, $\hat{\textbf{y}}_{union}^{\hat{T}_i}$, as  additional inputs to the RoIAligned features. We investigate two methods for embedding global context information: \textbf{1) Embedding in convolutional layers:} We embed $\hat{\textbf{y}}_{union}^{\hat{T}_i}$ to modulate features after each convolutional layer of the tubelet classifier, following the feature modulation method proposed in \cite{li2022noise, xia2021ct}. The embedded feature, obtained via a fully connected (FC) layer, was processed through multiple channel-attention blocks to condition feature map importance on the tubelet’s global spatial information. Channel-attention conditioning is applied by multiplying the channel feature vector, derived from a 2-D adaptive max pooling layer and two 1x1 convolutions, with the first half of the embedding layer’s feature vector, then adding the result to the second half. 
\textbf{2) Embedding in FC layer:}
Rather than modulating features within convolutional layers, we embed $\hat{\textbf{y}}_{union}^{\hat{T}_i}$ into the FC layer after temporal max-pooling (section \ref{tubelet_classifier}). Specifically, we use an FC layer to encode $\hat{\textbf{y}}_{union}^{\hat{T}_i}$ into a 1-D feature vector that matches the size of the temporally pooled image feature of the tubelet. This encoded non-image feature is then concatenated with the temporally pooled image feature of the tubelet and passed through another FC layer for tubelet classification. 

\subsection{Spatiotemporal tubelet classifier}
\label{tubelet_classifier}
As shown in \figref{fig:main}, the newly introduced tubelet classifier processes the entire tubelet by first encoding each instance (in red; $f_{i}$ from Eq. 2) in each frame and then temporally pooling the encoded instance-level features to form the tubelet-level feature, which is used for tubelet-level classification. Specifically, each instance ($R^{C \times {ROI}_{res}^2}$) is an RoIAligned feature extracted from the detector’s feature maps based on predicted bounding boxes within a tracklet. These are passed through a shared CNN encoder ($N$ convolutional blocks) to obtain encoded instance-level image features ($R^{C\times x^2}$). The features are then flattened and sent to a shared fully connected layer for instance-level classification, supverised by the instance-level class label $L_i^t=1$ (see section \ref{gen_video_tubelets}). Temporal pooling is applied using a max-pooling layer across all flattened instance features in the tubelet, forming a 1D tubelet feature (blue). This 1-D tubelet image feature is concatenated with the embedded non-image tubelet feature (orange), and the combined feature is fed into another fully connected layer for tubelet-level classification, supervised by $y_i$ from Eq. 4.

\section{Results}
\label{sec:majhead}
\subsection{Experimental settings}
\textbf{Dataset} The study dataset comprised 14,804 LUS videos acquired across 11 U.S. clinical sites from 828 patients suspected of consolidation (CON) and/or pleural effusion (PE). The dataset was divided into five cross-validation folds, with 6,682 videos used for testing across the five folds. Each video was annotated by three ultrasound fellowship-trained physicians for the presence or absence of CON and PE, and the final video classification was determined by majority vote. Frame-by-frame box annotations were done by ultrasound-trained researchers and reviewed by at least one physician.

\noindent \textbf{Implementation details} The YOLO-v5 detector was trained in a weakly semi-supervised manner per \cite{ouyang2023weakly}. The SORT tracker was set with a minimum track length of 5, a hit threshold of 3, and an IoU threshold of 0.5. The tubelet classifier was trained for both pathologies with  ${ROI}_{res}=32$ and $x=4$, doubling the number of feature maps with each downsampling. Training ran for 500 epochs with binary cross-entropy loss at both instance ($L_{ins}$) and tubelet ($L_{tubelet}$) levels, using Adam optimizer with learning rate of 1e-5.

\subsection{Comparison methods for video-level classification}
\label{ssec:subhead}
\tabref{tab:comparison} summarizes video classification performance, measured by mean AUROC across the five cross-validation folds, for several recently proposed state-of-the-art methods, alongside our approach. Comparing direct video classification using a CNN-LSTM \cite{shea2023deep} (\tabref{tab:comparison}, row 1) with the frame aggregation approach proposed in \cite{li2023weakly} (\tabref{tab:comparison}, row 2), we observed a 10.0\% AUROC improvement for PE classification and a 15.1\% improvement for CON. This suggests that CON classification requires finer texture discrimination, while PE classification performs well with good temporal modeling. 
Our method, which incorporates ROI feature maps into video tubelets and adds contextual embeddings, led to further AUROC improvements: for PE, there was a 1.1\% increase using embeddings in all convolutional layers (\tabref{tab:comparison}, row 5) and 1.4\% using embeddings in the final FC layer (\tabref{tab:comparison}, row 6), while for consolidation, these increases were 1.4\% and 2.3\%, respectively. The results indicate that, together with incorporating ROI feature maps, both context embedding methods lead to statistically significant improvements (p$\leq$0.05) over a strong detector-based video classification baseline \cite{li2023weakly}, with FC embeddings slightly outperforming convolutional layer embeddings. \figref{fig:tubeletClsExample} shows example predictions from our context-aware video classification framework.

\begin{table}[t!]
  \scriptsize
  \renewcommand{\arraystretch}{1.12}
  \setlength\tabcolsep{1.15pt}
  \caption{
  Comparisons with SOTA methods. Video-level AUROC: mean test AUROC on five cross-validation folds; p: p-value from two-sided paired t-test; TR: SORT tracking; ST-TubeCls: Spatiotemporal tubelet classifier; RoIA: RoI align on F4; EmbedAll: embed $\hat{\textbf{y}}_{union}^{\hat{T}_i}$ at all stages; EmbedLast: embed $\hat{\textbf{y}}_{union}^{\hat{T}_i}$ at last layer.}
  
  \begin{tabular}{lcc} 
  \toprule
   & \multicolumn{2}{c}{Video-level AUROC} \\
  &\textit{Pleural effusion} &\textit{Consolidation}  \\
  
  \hline
 {\textcolor{blue}{CNN-LSTM}\cite{shea2023deep}} & 83.0\% & 74.5\% \\
  \hline
{\textcolor{teal}{YOLO + SumConf} \cite{ouyang2023weakly}} &  { \centering 92.3\%  (p<0.001 vs \cite{shea2023deep})} & { \centering 90.4\% (p<0.001 vs \cite{shea2023deep})}  \\    

 {\textcolor{teal}{YOLO + TR + CNN-LSTM} \cite{li2023weakly}} &  { \centering 93.0\%  (p<0.001 vs \cite{shea2023deep})} & { \centering 89.6\% (p<0.001 vs \cite{shea2023deep})}  \\    
  \hline



 {YOLO + TR + ST-TubeCls} &  { \centering 93.7\%  (p<0.01 vs \cite{li2023weakly})} & { \centering 90.1\% (p=0.08 vs \cite{li2023weakly})}  \\    
  
  \multirow{2}{0.35\columnwidth}{YOLO + TR + ST-TubeCls + \\RoIA + EmbedAll} & \multirow{2}{0.3\columnwidth}{\centering 94.1\% \\ (p<0.01 vs \cite{li2023weakly})} & \multirow{2}{0.3\columnwidth}{\centering 91.0\% \\ (p<0.01 vs \cite{li2023weakly})} \\ \\
  
  \multirow{2}{0.35\columnwidth}{YOLO + TR + ST-TubeCls + \\RoIA + EmbedLast} & \multirow{2}{0.3\columnwidth}{ \centering \textbf{94.4}\% \\ (p<0.05 vs \cite{li2023weakly})} & \multirow{2}{0.3\columnwidth}{\centering \textbf{91.9}\% \\ (p<0.001 vs \cite{li2023weakly})}  \\ \\ 
  \bottomrule
  \end{tabular}
  \label{tab:comparison}
\end{table}



\subsection{Ablation study}
\noindent\textbf{Video tubelets with detector feature maps and contextual embeddings:} \tabref{tab:ablation} summarizes ablation studies evaluating the impact of introducing feature maps and contextual embeddings for tubelet classification. Compared to the baseline method excluding both components \cite{li2023weakly} (\tabref{tab:ablation}, row 1), our approach led to a 1\% improvement in AUROC for PE classification and a 1.2\% improvement for CON (\tabref{tab:comparison}, row 4).

\begin{table}[t!]
\centering
\scriptsize
\renewcommand{\arraystretch}{1.15}
\setlength\tabcolsep{2.35pt}
\caption{Ablation experiments; Video-level AUROC: test AUROC evaluated on a single cross-validation fold.}
\begin{tabular}{c c c c c c}
  \toprule

& & & & \multicolumn{2}{c}{Video-level (AUROC)} \\
Embed\newline Last & RoIA\newline @F4 & Params & FLOPs &\textit{Pleural effusion} &\textit{Consolidation}  \\ 

\hline
     &              & 1.6M & 36.8G & 90.9\% & 92.0\%  \\
          & \checkmark    & \textbf{0.4M} & \textbf{6.8G}  & 91.7\% & 92.6\%  \\
\checkmark &  & 1.6M & 36.8G & 91.9\% & 91.9\%  \\ 
\checkmark & \checkmark  & \textbf{0.4M} & \textbf{6.8G}  & \textbf{91.9}\% & \textbf{93.2}\%  \\
\hline
\end{tabular}
\label{tab:ablation}
\end{table}


\noindent\textbf{Detector feature maps:} 
Before introducing contextual embeddings, we first compared applying ROI align on the original image (\tabref{tab:ablation}, row 1) versus using ROI align on feature maps (\tabref{tab:ablation}, row 2). This resulted in a 0.8\% and 0.6\% AUROC improvement for PE and CON, respectively, along with a 6x reduction in FLOPs. This demonstrates a significant speed improvement and a modest performance gain by reusing feature representations learned through the strongly supervised detection loss. When only introducing  contextual embeddings without feature maps (\tabref{tab:ablation}, row 3), we observed a slight drop (0.1\%) for CON but a significant improvement (1\%) for PE. Combining both approaches (\tabref{tab:ablation}, row 4) led to an additional 1.3\% AUROC improvement for CON, indicating that the two methods can work synergistically. For completeness, we also investigated extracting feature maps from different detector layers (\tabref{tab:trend}).

\noindent\textbf{Contextual embeddings:}
After introducing feature maps, we compared video classification performance with versus without contextual embeddings, observing an AUROC increase of 0.2\% and 0.6\% for PE and CON classification, respectively.(\tabref{tab:ablation}, row 2 versus row 4).


\begin{table}[t!]
\centering
\scriptsize
\renewcommand{\arraystretch}{1.15}
\setlength\tabcolsep{4.35pt}
\caption{Detector feature extraction using RoI align. F*: feature maps from the final convolutional layer of the $*^{th}$ stage of the detector.
$\text{RoI}_{\text{res}}$: Feature map resolution after applying RoI align.}
\begin{tabular}{c c c c c}
  \toprule
& & & \multicolumn{2}{c}{Video-level (AUROC)} \\
& Params & $\text{RoI}_{\text{res}}$ & \textit{Pleural effusion} & \textit{Consolidation}  \\

\hline
Original image  & 1.6M & 128 & 91.9 \% & 91.9 \%  \\ 
F2  & 0.4M & 64 & 91.9\% & 92.4\%   \\ 
\rowcolor{mygray}
F4  & 0.5M & 32 & \textbf{91.9}\% &\textbf{93.2} \%   \\ 
F6  & 0.4M  & 16 & 90.7\% & 92.7\%   \\ 
F8  & 0.3M  & 8 & 90.9\% & 92.5\%    \\
\hline
\end{tabular}
\label{tab:trend}
\end{table}

\section{Conclusion}
\label{sec:Conclusions}
We presented a method for simultaneous frame detection and video classification that: (1) integrates spatial context embeddings to retain global information, (2) aggregates detector features into tubelets, and (3) employs an efficient spatiotemporal model for tubelet processing. Cross-validation and ablation studies demonstrated improved classification for two LUS pathologies compared to previous techniques. Future work will explore alternative ROI feature extraction methods \cite{gong2021temporal}, other detector architectures, and the use of this method for other ultrasound applications, such as cardiac imaging, where spatiotemporal analysis may be of beneficial.

\vfill
\pagebreak

\bibliographystyle{IEEEbib}
\bibliography{strings,refs}

\begin{thebibliography}{10}

\bibitem{li2023weakly}
Gary~Y Li, Li~Chen, Mohsen Zahiri, Naveen Balaraju, Shubham Patil, Courosh Mehanian, Cynthia Gregory, Kenton Gregory, Balasundar Raju, Jochen Kruecker, et~al.,
\newblock ``Weakly semi-supervised detector-based video classification with temporal context for lung ultrasound,''
\newblock in {\em Proceedings of the IEEE/CVF International Conference on Computer Vision}, 2023, pp. 2483--2492.

\bibitem{shea2023deep}
Daniel~E Shea, Sourabh Kulhare, Rachel Millin, Zohreh Laverriere, Courosh Mehanian, Charles~B Delahunt, Dipayan Banik, Xinliang Zheng, Meihua Zhu, Ye~Ji, et~al.,
\newblock ``Deep learning video classification of lung ultrasound features associated with pneumonia,''
\newblock in {\em Proceedings of the IEEE/CVF Conference on Computer Vision and Pattern Recognition}, 2023, pp. 3103--3112.

\bibitem{dastider2021integrated}
Ankan~Ghosh Dastider, Farhan Sadik, and Shaikh~Anowarul Fattah,
\newblock ``An integrated autoencoder-based hybrid cnn-lstm model for covid-19 severity prediction from lung ultrasound,''
\newblock {\em Computers in Biology and Medicine}, vol. 132, pp. 104296, 2021.

\bibitem{sharma2019spatio}
Harshita Sharma, Richard Droste, Pierre Chatelain, Lior Drukker, Aris~T Papageorghiou, and J~Alison Noble,
\newblock ``Spatio-temporal partitioning and description of full-length routine fetal anomaly ultrasound scans,''
\newblock in {\em 2019 IEEE 16th International Symposium on Biomedical Imaging (ISBI 2019)}. IEEE, 2019, pp. 987--990.

\bibitem{hwang2022cannot}
Sukjun Hwang, Miran Heo, Seoung~Wug Oh, and Seon~Joo Kim,
\newblock ``Cannot see the forest for the trees: Aggregating multiple viewpoints to better classify objects in videos,''
\newblock in {\em Proceedings of the IEEE/CVF Conference on Computer Vision and Pattern Recognition}, 2022, pp. 17052--17061.

\bibitem{kang2016object}
Kai Kang, Wanli Ouyang, Hongsheng Li, and Xiaogang Wang,
\newblock ``Object detection from video tubelets with convolutional neural networks,''
\newblock in {\em Proceedings of the IEEE conference on computer vision and pattern recognition}, 2016, pp. 817--825.

\bibitem{kang2017object}
Kai Kang, Hongsheng Li, Tong Xiao, Wanli Ouyang, Junjie Yan, Xihui Liu, and Xiaogang Wang,
\newblock ``Object detection in videos with tubelet proposal networks,''
\newblock in {\em Proceedings of the IEEE conference on computer vision and pattern recognition}, 2017, pp. 727--735.

\bibitem{yue2015beyond}
Joe Yue-Hei~Ng, Matthew Hausknecht, Sudheendra Vijayanarasimhan, Oriol Vinyals, Rajat Monga, and George Toderici,
\newblock ``Beyond short snippets: Deep networks for video classification,''
\newblock in {\em Proceedings of the IEEE conference on computer vision and pattern recognition}, 2015, pp. 4694--4702.

\bibitem{lin2022new}
Zhi Lin, Junhao Lin, Lei Zhu, Huazhu Fu, Jing Qin, and Liansheng Wang,
\newblock ``A new dataset and a baseline model for breast lesion detection in ultrasound videos,''
\newblock in {\em International Conference on Medical Image Computing and Computer-Assisted Intervention}. Springer, 2022, pp. 614--623.

\bibitem{roy2020deep}
Subhankar Roy, Willi Menapace, Sebastiaan Oei, Ben Luijten, Enrico Fini, Cristiano Saltori, Iris Huijben, Nishith Chennakeshava, Federico Mento, Alessandro Sentelli, et~al.,
\newblock ``Deep learning for classification and localization of covid-19 markers in point-of-care lung ultrasound,''
\newblock {\em IEEE transactions on medical imaging}, vol. 39, no. 8, pp. 2676--2687, 2020.

\bibitem{han2016seq}
Wei Han, Pooya Khorrami, Tom~Le Paine, Prajit Ramachandran, Mohammad Babaeizadeh, Honghui Shi, Jianan Li, Shuicheng Yan, and Thomas~S Huang,
\newblock ``Seq-nms for video object detection,''
\newblock {\em arXiv preprint arXiv:1602.08465}, 2016.

\bibitem{he2017mask}
Kaiming He, Georgia Gkioxari, Piotr Doll{\'a}r, and Ross Girshick,
\newblock ``Mask r-cnn,''
\newblock in {\em Proceedings of the IEEE international conference on computer vision}, 2017, pp. 2961--2969.

\bibitem{bewley2016simple}
Alex Bewley, Zongyuan Ge, Lionel Ott, Fabio Ramos, and Ben Upcroft,
\newblock ``Simple online and realtime tracking,''
\newblock in {\em 2016 IEEE international conference on image processing (ICIP)}. IEEE, 2016, pp. 3464--3468.

\bibitem{li2022noise}
Ye~Li, Jianan Cui, Junyu Chen, Guodong Zeng, Scott Wollenweber, Floris Jansen, Se-In Jang, Kyungsang Kim, Kuang Gong, and Quanzheng Li,
\newblock ``A noise-level-aware framework for pet image denoising,''
\newblock in {\em International Workshop on Machine Learning for Medical Image Reconstruction}. Springer, 2022, pp. 75--83.

\bibitem{xia2021ct}
Wenjun Xia, Zexin Lu, Yongqiang Huang, Yan Liu, Hu~Chen, Jiliu Zhou, and Yi~Zhang,
\newblock ``Ct reconstruction with pdf: Parameter-dependent framework for data from multiple geometries and dose levels,''
\newblock {\em IEEE Transactions on Medical Imaging}, vol. 40, no. 11, pp. 3065--3076, 2021.

\bibitem{ouyang2023weakly}
Jiahong Ouyang, Li~Chen, Gary~Y Li, Naveen Balaraju, Shubham Patil, Courosh Mehanian, Sourabh Kulhare, Rachel Millin, Kenton~W Gregory, Cynthia~R Gregory, et~al.,
\newblock ``Weakly semi-supervised detection in lung ultrasound videos,''
\newblock in {\em International conference on information processing in medical imaging}. Springer, 2023, pp. 195--207.

\bibitem{gong2021temporal}
Tao Gong, Kai Chen, Xinjiang Wang, Qi~Chu, Feng Zhu, Dahua Lin, Nenghai Yu, and Huamin Feng,
\newblock ``Temporal roi align for video object recognition,''
\newblock in {\em Proceedings of the AAAI Conference on Artificial Intelligence}, 2021, vol.~35, pp. 1442--1450.

\end{thebibliography}

\end{document}